%% file: main.tex
\documentclass[11pt,letterpaper]{article}
\usepackage{liveaigc_report}

\input{math_commands.tex}

\usepackage{mathtools,xspace}
\usepackage{tikz}
\usetikzlibrary{arrows.meta,backgrounds,calc,fit,positioning}

\definecolor{lowres}{HTML}{DDEBFF}
\definecolor{highres}{HTML}{FFE3D6}
\definecolor{memory}{HTML}{E7E0F5}
\definecolor{darkblue}{HTML}{245A8D}
\definecolor{darkorange}{HTML}{B85723}
\newcommand{\method}{\textsc{TaoMate-HD}\xspace}
\newcommand{\continuationbaseline}{\textit{w/o asymmetric distillation}\xspace}
\newcommand{\lowcanvas}{480$\times$864}
\newcommand{\highcanvas}{768$\times$1376}
\input{metadata}

\begin{document}
\maketitle
\input{sections/0_abstract}
\input{figures/teaser}
\input{sections/1_introduction}
\FloatBarrier
\input{sections/2_related_work}
\FloatBarrier
\input{sections/3_method}
\FloatBarrier
\input{sections/4_experiments}

\input{sections/5_conclusion}
\label{paper:conclusion_end}

\clearpage
\label{paper:references_start}
\bibliographystyle{unsrtnat}
\bibliography{iclr2027_conference}

\end{document}

%% file: math_commands.tex
\usepackage{amsmath,amsfonts,bm}

\def\eqref#1{equation~\ref{#1}}

\def\1{\bm{1}}

\DeclareMathAlphabet{\mathsfit}{\encodingdefault}{\sfdefault}{m}{sl}
\SetMathAlphabet{\mathsfit}{bold}{\encodingdefault}{\sfdefault}{bx}{n}



%% file: metadata.tex
\title{TaoMate-HD:\\Progressive-Resolution Autoregressive\\Audio-Video Generation}
\reportauthors{Qijun Gan\textsuperscript{1,*}, Chenwei Zhang\textsuperscript{1,*}, Xinyang Li\textsuperscript{1},
Meiguang Jin\textsuperscript{1,\textdagger}, Junfeng Ma\textsuperscript{1}, Shen Qiu\textsuperscript{2}}
\authornote{\textsuperscript{*}Equal contribution.\quad \textsuperscript{\textdagger}Corresponding author.}
\affiliation{\textsuperscript{1}Alibaba Group\qquad \textsuperscript{2}Nanjing University\\[0.3em]
\texttt{\{ganqijun.gqj, meiguang.jmg\}@taobao.com}}
\teamname{}
\shorttitle{TaoMate-HD}
\reportlabel{Technical Report}
\reportdate{October 2026}

\leftlogo[height=0.9cm]{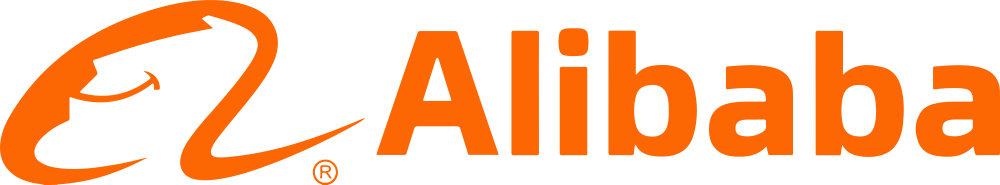}
\rightlogo[height=0.9cm,trim=436bp 638bp 306bp 480bp,clip]{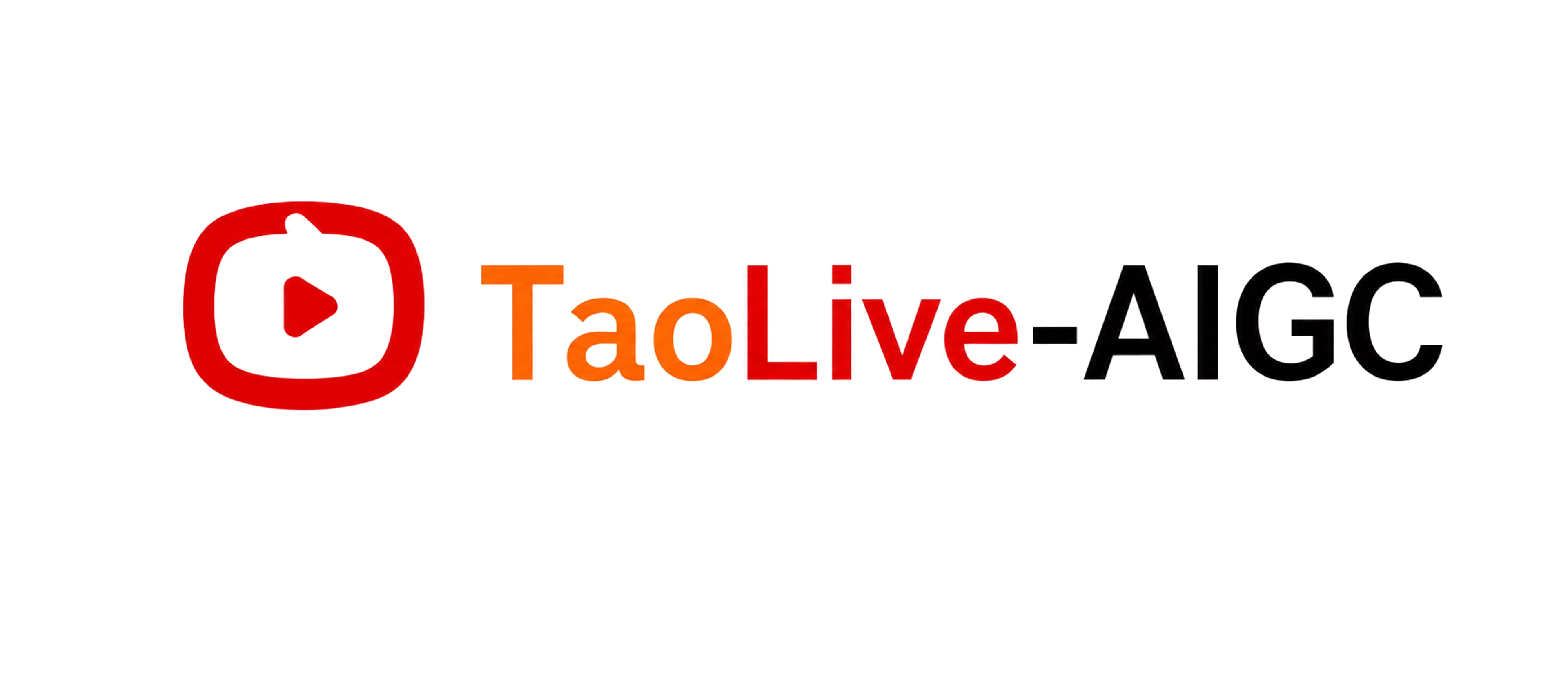}

\hypersetup{
  pdftitle={TaoMate-HD: Progressive-Resolution Autoregressive Audio-Video Generation},
  pdfauthor={Qijun Gan, Chenwei Zhang, Xinyang Li, Meiguang Jin, Junfeng Ma, Shen Qiu},
  pdfsubject={Progressive-resolution autoregressive audio-video generation},
  pdfkeywords={audio-video generation, progressive resolution, autoregressive generation, asymmetric distillation, stage-parallel inference}
}

%% file: sections/0_abstract.tex
\begin{abstract}
Streaming digital humans produce detailed video while continuing speech as new text arrives.
High-resolution denoising is costly, and text updates can cause repeated or interrupted speech.
We present \method, an autoregressive model for high-resolution audio-video streaming.
For each short segment, or chunk, two denoising steps establish coarse video structure at low resolution.
A final step refines the video to high resolution, with audio generated throughout the same trajectory.
Each stage passes its audiovisual predictions to the same stage of the next chunk as history.
Across text updates, the model retains this history to infer what has already been spoken.
Continuation training utilizes asymmetric text conditioning, with separate text windows for generation and distillation.
Experiments show a $1.6\times$ speedup over direct high-resolution generation with the same model and hardware.
Optimized stage-parallel deployment provides a further $1.9\times$ overall speedup over serial progressive generation.
Our continuation training reduces character error rate by 11.18 percentage points relative to the baseline.
The model also supports hour-long audiovisual generation. Our project page is \url{https://taoliveaigc.github.io/TaoMate-HD}.
\end{abstract}

%% file: figures/teaser.tex
\begin{figure*}[t]
\centering
\begin{tikzpicture}[
  x=1mm,y=-1mm,
  font=\sffamily\fontsize{7}{8}\selectfont,
  tile/.style={draw=black!25,line width=.35pt,rounded corners=.6pt,
    fill=highres,minimum width=9mm,minimum height=5.5mm,inner sep=0pt},
  flow/.style={-{Latex[length=1mm,width=.7mm]},line width=.45pt,draw=black!60},
  labeltext/.style={inner sep=0pt,align=center},
  heading/.style={labeltext,font=\sffamily\fontsize{8}{9}\selectfont\bfseries}
]
\path[use as bounding box] (0,0) rectangle (139.7,22);
\node[heading] at (22,1.8) {MiniMax-H3};
\node[heading] at (68.5,1.8) {TaoMate};
\node[heading,text=darkblue] at (116,1.8) {TaoMate-HD};

\foreach \offset [count=\methodindex] in {0}{
  \node[tile] (a\methodindex) at ({7+\offset},8) {H};
  \node[tile] (b\methodindex) at ({20+\offset},8) {H};
  \node[labeltext] at ({29+\offset},8) {$\cdots$};
  \node[tile] (c\methodindex) at ({37.5+\offset},8) {H};
  \draw[flow] (a\methodindex.east) -- (b\methodindex.west);
  \node[labeltext,text=black!65] at ({22+\offset},17.5) {Current-window text};
  \node[labeltext,text=black!65] at ({22+\offset},20.5) {Previous frame $\rightarrow$ next clip};
}
\node[labeltext] at (22,13.6) {49 denoising steps / clip};
\node[tile] (ta) at (53.5,8) {H};
\node[tile] (tb) at (68.5,8) {H};
\node[tile] (tc) at (83.5,8) {H};
\draw[flow] (ta.east) -- (tb.west);
\draw[flow] (tb.east) -- (tc.west);
\node[labeltext] at (68.5,13.6) {3 denoising steps / chunk};
\node[labeltext,text=black!65] at (68.5,17.5) {Fixed-resolution generation};
\node[labeltext,text=black!65] at (68.5,20.5) {Persistent AV memory $\rightarrow$ next chunk};

\node[tile,fill=lowres,minimum width=6mm,minimum height=3.6mm] (l0) at (100,8) {L};
\node[tile,fill=lowres,minimum width=6mm,minimum height=3.6mm] (l16) at (112,8) {L};
\node[tile] (h33) at (129,8) {H};
\draw[flow] (l0.east) -- (l16.west);
\draw[flow] (l16.east) -- (h33.west);
\node[labeltext] at (116,13.6) {3 denoising steps / chunk};
\node[labeltext,text=darkblue] at (116,17.5) {Asymmetric text conditioning};
\node[labeltext,text=darkblue] at (116,20.5) {Stage-local AV KV $\rightarrow$ next chunk};
\end{tikzpicture}
\par\vspace{2mm}
\includegraphics[width=\linewidth]{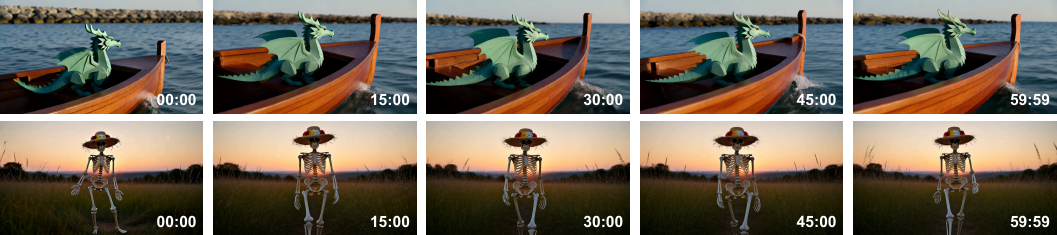}
\par\captionsetup{skip=3pt}
\caption{\textbf{Progressive audiovisual generation sustained over an hour.}
Top: MiniMax-H3 denoises high-resolution clips and passes a frame to the next clip, while TaoMate uses few-step fixed-resolution generation with persistent audiovisual memory.
\method refines each chunk through two low-resolution evaluations and one high-resolution evaluation, passing each stage's audiovisual prediction to the next chunk.
Retained history supports speech continuation as text windows advance.
L/H denote low/high video grids.
Bottom: full frames from two 60-minute, $1376\times768$ audiovisual streams at the labeled times (minutes:seconds).
The sampled frames preserve character appearance and scene composition across the hour.}
\label{fig:teaser}
\end{figure*}

%% file: sections/1_introduction.tex
\section{Introduction}
\label{sec:introduction}

Interactive digital humans need to sustain speech and visual motion as new dialogue arrives.
Joint audio-video models generate both modalities from text \citep{ltx2_2026,javisditpp2026,ovi2025,mova2026}.
Recent streaming systems extend this process over successive short segments, or chunks \citep{omniforcing2026,taomate2026,mao2026lynnrealomninativemultimodalvideo}.
These streams must preserve fine details in faces, hands, and foreground objects while keeping speech continuous.
Meeting both requirements is difficult: high-resolution synthesis is expensive, and text updates can disrupt ongoing speech.

Direct high-resolution generation processes a dense video-token grid at every denoising step.
Even early steps use this grid when they mainly establish scene layout and motion.
Few-step distillation reduces the number of steps \citep{causvid2025,selfforcing2025}.
Progressive-resolution methods further reduce spatial computation by using coarser grids early in denoising \citep{pyramidal2025,pyramidalwan2026}.
However, faster processing of one chunk does not remove the wait between chunks.
If a chunk needs the previous chunk's final output as history, it must wait for all refinement to finish.
Stage-parallel inference overlaps denoising across chunks \citep{zou2026hiar}, but requires usable audiovisual context before that final output.

Speech continuity poses a related challenge when text changes during generation.
Streaming models retain audiovisual context across chunks \citep{omniforcing2026,taomate2026,mao2026lynnrealomninativemultimodalvideo}.
To continue speech, they must relate incoming text to the words already spoken.
Text-window boundaries need not coincide with sentence endings, so an update may arrive mid-sentence and include words already spoken.
Restarting from the first word repeats content, while stopping at the window boundary interrupts speech.
The model therefore needs recent audio to distinguish spoken context from upcoming dialogue.

We present \method, an autoregressive model for high-resolution audio-video streaming (Figure~\ref{fig:teaser}).
Its central idea is to reuse audiovisual predictions before final spatial refinement is complete.
Each chunk first undergoes two denoising steps at low resolution to establish coarse video structure.
One final step refines the video to high resolution, while audio is generated throughout the joint trajectory.
Each stage estimates clean audio and video without waiting for the remaining denoising steps.
These estimates provide history for the same stage of the next chunk.
The next chunk can thus begin coarse generation before the previous chunk finishes high-resolution refinement.

When the text window changes, we retain the audiovisual history and use overlapping text to provide continuation context.
The model infers speech progress from the retained audio, without an explicit text pointer or online speech recognizer.
We train this behavior on the model's own generated sequences, including text-window changes and speech completion.
Training uses asymmetric text conditioning to separate generation from sequence-level supervision.
The generator receives upcoming dialogue, while distillation uses the opening text context to evaluate the full audiovisual sequence.

Experiments show that progressive generation is $1.60\times$ faster than direct high-resolution generation with the same model and hardware.
Compared with the baseline, our continuation training reduces character error rate from 30.75\% to 19.57\%.
We also demonstrate hour-long audiovisual generation.

\begin{samepage}
Our contributions are threefold:
\begin{list}{$\bullet$}{\setlength{\leftmargin}{\leftmargini}\setlength{\topsep}{2pt}\setlength{\partopsep}{0pt}}
\setlength{\itemsep}{1pt}
\setlength{\parskip}{0pt}
\item A progressive audio-video generator that concentrates denoising on a coarse grid. Stage-local audiovisual history enables overlapping generation across chunks.
\item Continuation training with asymmetric text conditioning and student-generated history to support speech across overlapping text windows.
\item Hour-long high-resolution audiovisual demonstrations and quantitative comparisons of spatial efficiency, deployment throughput, and transcript accuracy.
\end{list}
\end{samepage}

%% file: sections/2_related_work.tex
\section{Related Work}
\label{sec:related}

\paragraph{Few-step autoregressive audio-video generation.}
Joint diffusion and flow models couple visual motion with speech and other acoustic content \citep{ltx2_2026,ovi2025,javisditpp2026,mova2026}.
Progressive distillation, consistency training, and distribution matching reduce sampling to a small number of evaluations \citep{progressive_distill2022,consistency_models2023,dmd2023,dmd2_2024}.
For streaming generation, these few-step trajectories must also learn to condition on previously generated content.
CausVid uses distribution matching to distill a bidirectional teacher into a causal generator, with teacher ODE trajectories for initialization \citep{causvid2025}.
Self Forcing trains on student-generated histories and samples denoising exits for gradient computation \citep{selfforcing2025}.
Recent AV systems combine such rollouts with multimodal objectives and streaming inference \citep{omniforcing2026,mutualforcing2026,vorchstreamer2026, hallolive2026,mainecoon2026}.
LynnReal-Omni provides Standard and Flash joint-AV generators and extends generation through bounded latent history and boundary overlap \citep{mao2026lynnrealomninativemultimodalvideo}.
We likewise train on student-generated histories, but allow the generator's text window to advance while the distillation scores evaluate the sequence from its opening context.

\paragraph{Progressive spatial computation.}
Few-step sampling still pays for a dense grid at each evaluation.
Progressive spatial methods reduce this cost by varying resolution along the trajectory.
Matryoshka diffusion and pyramidal flow use multiple spatial scales within generation \citep{matryoshka2024,pyramidal2025}.
PyramidalWan first adapts the original-architecture teacher to multiple resolutions through flow matching, then studies distribution matching with this teacher and a refitted pyramidal teacher \citep{pyramidalwan2026}.
Its few-step schedule concentrates target-resolution computation in a single evaluation.
Spectral schedules and adaptive patch granularity likewise vary spatial computation with noise level \citep{spectralprogressive2026,msforcing2026}.
Video DeltaNet instead changes video-to-video attention through local Softmax, boundary anchors, and bidirectional linear memory, retaining Softmax interactions with text and audio \citep{xi2026videodeltanet}.
Causal super-resolution methods such as Ultra Flash take another approach, refining the output of a separate generator \citep{ultraflash2026}.
We change resolution within the joint audiovisual trajectory rather than after generation, and make the intermediate audiovisual predictions available as history for subsequent chunks.

\paragraph{Temporal memory and denoising parallelism.}
Across chunks, efficiency also depends on when history becomes available.
Long-video generators retain context through windows of cached features, recurrent state, selected history, or persistent anchors \citep{streamingt2v2025,longlive2026,videossm2025,adastate2026, slotmemory2026,tethercache2026}.
TaoMate separates an immutable visual anchor and adaptive audiovisual memory from the active KV cache, using memory attention and reference-aware modulation for long-form consistency \citep{taomate2026}.
Its stage-parallel inference operates at a fixed spatial resolution.
Ripple also extends recurrent temporal state to joint audio-video generation \citep{ripple2026}.
Stage-parallel inference organizes chunks and denoising stages into an anti-diagonal pipeline and fuses context writing with the next chunk's denoising \citep{zou2026hiar}.
The stage-specific history in that formulation is represented at the stage's output noise level.
We use this execution organization with paired clean audiovisual predictions on each stage's native grid.
The clean prediction provides a common history representation even though the query noise level and spatial grid differ between stages.

%% file: sections/3_method.tex
\section{Method}
\label{sec:method}

\method is an autoregressive audio-video model that uses coarse-to-fine denoising and stage-local KV history to generate a continuous stream (Figure~\ref{fig:method_pipeline}).
Each chunk develops on a coarse video grid before high-resolution refinement, while intermediate audiovisual predictions provide context for subsequent chunks.
Training on the model's own rollouts with asymmetric text conditioning supports speech continuation as the dialogue advances.

\input{figures/method_pipeline}

\subsection{Progressive generation}
\label{sec:progressive}
\label{sec:trajectory}

Direct high-resolution generation repeats full-grid attention at every denoising step, including early steps that establish scene layout and large-scale motion.
Progressive generation assigns this structural synthesis to a coarse grid and defers fine spatial refinement to the output grid.
For an $N$-step trajectory, we perform the first $N-1$ evaluations on a low-resolution video grid $L$ and the final evaluation on a high-resolution grid $H$.
Audio retains the same latent geometry throughout, so video refinement and audio generation remain part of one joint trajectory.

Let $u_{b,\ell}^m$ denote the noisy latent of modality $m\in\{v,a\}$ in chunk $b$ at stage $\ell\in\{1,\ldots,N\}$.
Each stage consists of one generator evaluation and a denoising update, with decreasing noise levels $\sigma_\ell^m$ and terminal level $\sigma_{N+1}^m=0$.
Conditioned on text $c_b^G$ and preceding audiovisual history $\mathcal H_{b,\ell}$, the generator predicts a denoising velocity $\widehat v_{b,\ell}^m$.
Under the flow parameterization $u_\sigma=(1-\sigma)x_{\mathrm{clean}}+\sigma\epsilon$ \citep{liu2023rectifiedflow}, this velocity estimates $x_{\mathrm{clean}}-\epsilon$.
On the current grid, it yields both an Euler update and a clean prediction:
\begin{equation}
\begin{aligned}
\widetilde u_{b,\ell+1}^m
 &=u_{b,\ell}^m+(\sigma_\ell^m-\sigma_{\ell+1}^m)\widehat v_{b,\ell}^m,\\
\widehat x_{b,\ell}^m
 &=u_{b,\ell}^m+\sigma_\ell^m\widehat v_{b,\ell}^m.
\end{aligned}
\label{eq:solver}
\end{equation}
The clean prediction estimates the zero-noise endpoint directly from the current velocity, without executing the remaining stages.
It supplies history to the next chunk, while the Euler update advances the current chunk.
At the final stage, the two coincide because the next noise level is zero.

Before the final evaluation, the video state must move from $L$ to $H$ with its signal and residual noise intact.
Direct interpolation correlates the residual noise, whereas the high-resolution trajectory requires standard Gaussian noise.
We therefore couple high-grid noise $\epsilon_H\sim\mathcal N(0,I)$ with low-grid noise $\epsilon_L=P\epsilon_H$, where $PP^\top=I$ keeps the low-grid noise standard Gaussian.
Using bilinear interpolation $U$, the transition at noise level $\sigma$ is
\begin{equation}
\Lambda_\sigma(u_L)
 =U(u_L)+\sigma\bigl[\epsilon_H-U(\epsilon_L)\bigr].
\label{eq:lift}
\end{equation}
The correction replaces interpolated noise with its coupled high-grid counterpart, preserving the signal/noise decomposition.
The next stage takes the Euler-updated state, except at the final-stage entrance, where $u_{b,N}^v=\Lambda_{\sigma_N^v}(\widetilde u_{b,N}^v)$.
Audio passes through unchanged at this spatial transition.
Concentrating the earlier evaluations on $L$ reduces denoising computation at high resolution.
We instantiate this schedule with $N=3$, using grids $L$ (\lowcanvas{}) and $H$ (\highcanvas{}).

\subsection{Stage-local KV history}
\label{sec:memory}
\label{sec:streaming}
\label{sec:pipeline}

Progressive generation reduces computation within a chunk, but overlapping chunks also requires history before final refinement finishes.
We therefore give each stage its own key/value (KV) history bank, built from the preceding chunk's same-stage video and audio predictions.
The clean estimates in Equation~\ref{eq:solver} are encoded at zero noise ($\sigma=0$) on that stage's native spatial grid.
This encoding turns an intermediate endpoint estimate into reusable context without waiting for later denoising stages.
The coarse-stage banks remain separate because they contain predictions at different refinement levels, while the final-stage bank stores the high-resolution endpoint.

Let $\mathcal T_{b,\ell}$ denote stage $\ell\in\{1,\ldots,N\}$ of chunk $b$.
It requires the preceding stage of its own chunk and the same stage of the preceding chunk:
\begin{equation}
\mathcal T_{b,\ell}\ \leftarrow\
 \{\mathcal T_{b,\ell-1},\mathcal T_{b-1,\ell}\}.
\label{eq:pipeline_dependency}
\end{equation}
These dependencies permit stage-parallel inference \citep{zou2026hiar}: later chunks can occupy coarse stages while an earlier chunk undergoes final refinement.
Each stage keeps its KV history local and passes current-chunk audiovisual latents downstream.
Within a text window, coarse stages fuse encoding of the preceding chunk's clean estimate with denoising of the current noisy chunk \citep{zou2026hiar}.
A causal mask lets the noisy chunk read that estimate and earlier history, while the context chunk reads prior history and itself.

For long-stream inference, each bank retains the initial chunk's video as a visual anchor and the two most recent chunks' paired video and audio.
Initial audio remains available while it lies in this recent window.
The fixed-capacity banks preserve appearance and recent speech context without growing with stream duration.
KV and cached text projections are stored separately.
A text-window change refreshes the latter, leaving retained KV encoded under their original conditions.
Before the switch, any pending clean prediction is committed at noise using the old window that generated it.
Algorithm~\ref{alg:kv_window_update} summarizes this update for each stage.
\begin{algorithm}[H]
\caption{KV update at a text-window boundary}
\label{alg:kv_window_update}
\small
\begin{algorithmic}[1]
\State Commit any pending clean estimate under $c_{\mathrm{old}}^G$.
\State Retain initial-video KV and the two most recent AV chunks.
\State Activate $c_{\mathrm{new}}^G$ and refresh cached text projections.
\State Preserve retained KV and their temporal positions.
\State Denoise the next chunk with $c_{\mathrm{new}}^G$ and this history.
\end{algorithmic}
\end{algorithm}
Thus new text controls future synthesis without re-encoding the generated past.
Spatial coordinates remain native to each grid, and temporal positions continue across windows.
Training uses the same update order but retains the full causal prefix of its short rollouts.

\subsection{Asymmetric distillation for continuation}
\label{sec:training}
\label{sec:asymmetric}

Speech continuation requires relating new dialogue to the retained audio history.
We use overlapping text windows because updates can arrive mid-sentence.
Let $A$ be earlier text, $B$ the shared span, and $C$ new dialogue.
The generator moves from the initial window $W_1=A+B$ to $W_2=B+C$ while retaining history, using $B$ as context rather than restarting it.
At inference, audio history and text indicate speech progress without ASR, text pointers, or alignment modules.

We train this behavior on student-generated rollouts \citep{selfforcing2025}.
The student executes all $N$ stages, updates its own history banks, and switches text at a sampled chunk boundary.
To construct the windows, we extend source transcripts with same-topic speech while preserving the original wording, language, and speaker assignments.
Source speaking rate maps nominal time intervals to character or word spans.
The initial window $A+B$ covers the entire nominal rollout horizon even when the switch occurs earlier.

Distribution-matching distillation (DMD) supervises these rollouts by comparing a frozen teacher score with a learned score of the student's distribution \citep{dmd2023,dmd2_2024}.
The generator needs $W_2$ to continue the next chunk, whereas both scores evaluate the concatenated sequence from its beginning.
The learned \emph{fake score} therefore retains $W_1$: it includes the opening words absent from $W_2$ and the extended dialogue conditioning the student's rollout.
The frozen \emph{real score} instead uses the original source transcript prefix through the rollout duration.
This ties the teacher reference to the source segment, while the appended dialogue serves as generation lookahead.
Scene descriptions, speaker attributes, and audiovisual alignment instructions are shared across all three roles during training.

The generator uses \(W_1\) before the switch and \(W_2\) afterward, with ellipses marking omitted words:
\par\smallskip
\begingroup
\small
\setlength{\tabcolsep}{3pt}
\noindent
\begin{tabularx}{\linewidth}{@{}>{\raggedright\arraybackslash}p{26mm}>{\raggedright\arraybackslash}X@{}}
\textbf{Generator}\newline($W_2$) & \emph{\textbf{a set} of free templates you can download and use right away. These templates are designed to \ldots\ }\\[2pt]
\textbf{Fake}\newline($W_1$) & \emph{this into my own post-processing workflow. \ldots\ \textbf{a set} of free templates you can download and use right away. These templates are designed}\\[2pt]
\textbf{Real}\newline(source prefix) & \emph{this into my own post-processing workflow. \ldots\ I'll even give you \textbf{a set}}\\
\end{tabularx}
\endgroup
\par\smallskip
Both scores receive the same audiovisual sequence, corruption noise, and noise levels, so their difference defines the DMD update at a common noisy state.
The fake score learns from detached student sequences by velocity regression with an independent noise draw.

To supervise every refinement level, we sample one stage per rollout and concatenate its clean predictions across chunks.
The selected stage is shared across chunks so the sequence has a consistent refinement level.
For a coarse stage, bilinear interpolation enlarges the clean video latents to $H$, leaving time, channels, and audio unchanged.
This gives both scores a common high-resolution \emph{supervision canvas}.
The student continues to denoise on each stage's native grid.
Resizing the predicted clean video for scoring lets the high-resolution teacher supervise coarse stages without changing their computational resolution.
All $N$ stages still execute for every chunk, irrespective of the stage selected for supervision.

We first construct the rollout without gradients, then replay the selected evaluation one chunk at a time.
Each replayed prediction replaces its region in the otherwise detached canvas, and a chunk mask restricts the update to that region.
The scores retain full-sequence context, while gradients flow through the active prediction and its spatial interpolation, excluding other chunks and cached history.
An auxiliary teacher target supports final-stage detail refinement.
For a sampled noninitial chunk, the frozen base model starts from the student's lifted state, conditions on preceding final clean audiovisual chunks, and denoises to an endpoint.
We regress the student's final video prediction toward this target, complementing the joint audiovisual DMD objective.

Training also includes speech-completion examples so the model learns when to stop speaking as well as how to continue.
We select source clips whose endpoint is confirmed offline by the transcript, timestamped ASR, and voice activity detection \citep{silero2024vad}, and extend the rollout beyond that endpoint.
Here $A+B$ is the full original transcript: the generator first receives $A+B$, then its final span $B$, with no added $C$.
Both scores receive the full transcript and an instruction to finish the unspoken remainder and then produce natural nonspeech behavior.
Audio history indicates whether words remain to be spoken or the utterance has ended.

%% file: figures/method_pipeline.tex
\begin{figure*}[t]
\centering
\begingroup
\definecolor{figlow}{HTML}{1B5B88}
\definecolor{fighigh}{HTML}{9E481D}
\definecolor{figmem}{HTML}{5C428C}
\definecolor{figink}{HTML}{000000}
\definecolor{figmuted}{HTML}{000000}
\scalebox{1.05}{%
\begin{tikzpicture}[
  x=1mm,y=-1mm,
  font=\sffamily\fontsize{7.5}{8.5}\selectfont,text=figink,
  stage/.style={draw=#1!75,line width=.5pt,rounded corners=1.2mm,
    fill=#1!16,minimum width=34mm,minimum height=28mm,inner sep=0pt},
  bank/.style={draw=figmem!75,line width=.4pt,rounded corners=1.1mm,
    fill=figmem!23,minimum width=32mm,minimum height=5.6mm,inner sep=1pt,align=center},
  next/.style={draw=figlow!70,dash pattern=on 1.5pt off 1.2pt,line width=.4pt,
    fill=figlow!9,rounded corners=.8mm,minimum width=32mm,minimum height=4.6mm,inner sep=1pt,align=center},
  box/.style={draw=#1!75,line width=.45pt,rounded corners=1.1mm,
    fill=#1!20,minimum height=7mm,inner sep=1pt,align=center},
  tag/.style={draw=none,rounded corners=.7mm,fill=#1!20,text=#1,
    minimum height=4.5mm,inner xsep=1.2mm,inner ysep=.6mm,align=center,
    font=\sffamily\fontsize{6.7}{7.5}\selectfont},
  flow/.style={-{Latex[length=1.4mm,width=.9mm]},line width=.65pt,draw=black},
  history/.style={-{Latex[length=1.3mm,width=.85mm]},line width=.65pt,draw=figmem},
  condition/.style={-{Latex[length=1.1mm,width=.7mm]},line width=.4pt,draw=black!65},
  heading/.style={font=\sffamily\fontsize{8}{9}\selectfont\bfseries,anchor=west},
  labeltext/.style={inner sep=0pt,align=center},
  smalltext/.style={labeltext,font=\sffamily\fontsize{6.7}{7.4}\selectfont}
]
\path[use as bounding box] (0,-.5) rectangle (139.7,82);

\node[tag=figmuted,fill=black!13,minimum width=61mm,minimum height=4.8mm] (text) at (69,2.5)
  {Scene + dialogue condition $c_b^G$};
\draw[condition] (text.south) -- (69,6.8) -- (21,6.8) -- (21,9);
\draw[condition] (text.south) -- (69,9);
\draw[condition] (text.south) -- (69,6.8) -- (118,6.8) -- (118,9);

\node[stage=figlow] (low0) at (21,23) {};
\node[stage=figlow] (low16) at (69,23) {};
\node[stage=fighigh] (high33) at (118,23) {};
\node[labeltext,text=figlow] at (21,12) {\textbf{Stage 1}};
\node[labeltext,text=figlow] at (69,12) {\textbf{Stage 2}};
\node[labeltext,text=fighigh] at (118,12) {\textbf{Stage 3}};

\foreach \xx/\pixeloffset in {21/0,69/558,118/1115}{
  \begin{scope}
    \clip[rounded corners=.6mm] ({\xx-14},14.4) rectangle ({\xx+14},30.9);
    \node[anchor=north west,inner sep=0pt]
      at ({\xx-14},{14.4-\pixeloffset/940*28})
      {\includegraphics[width=28mm]{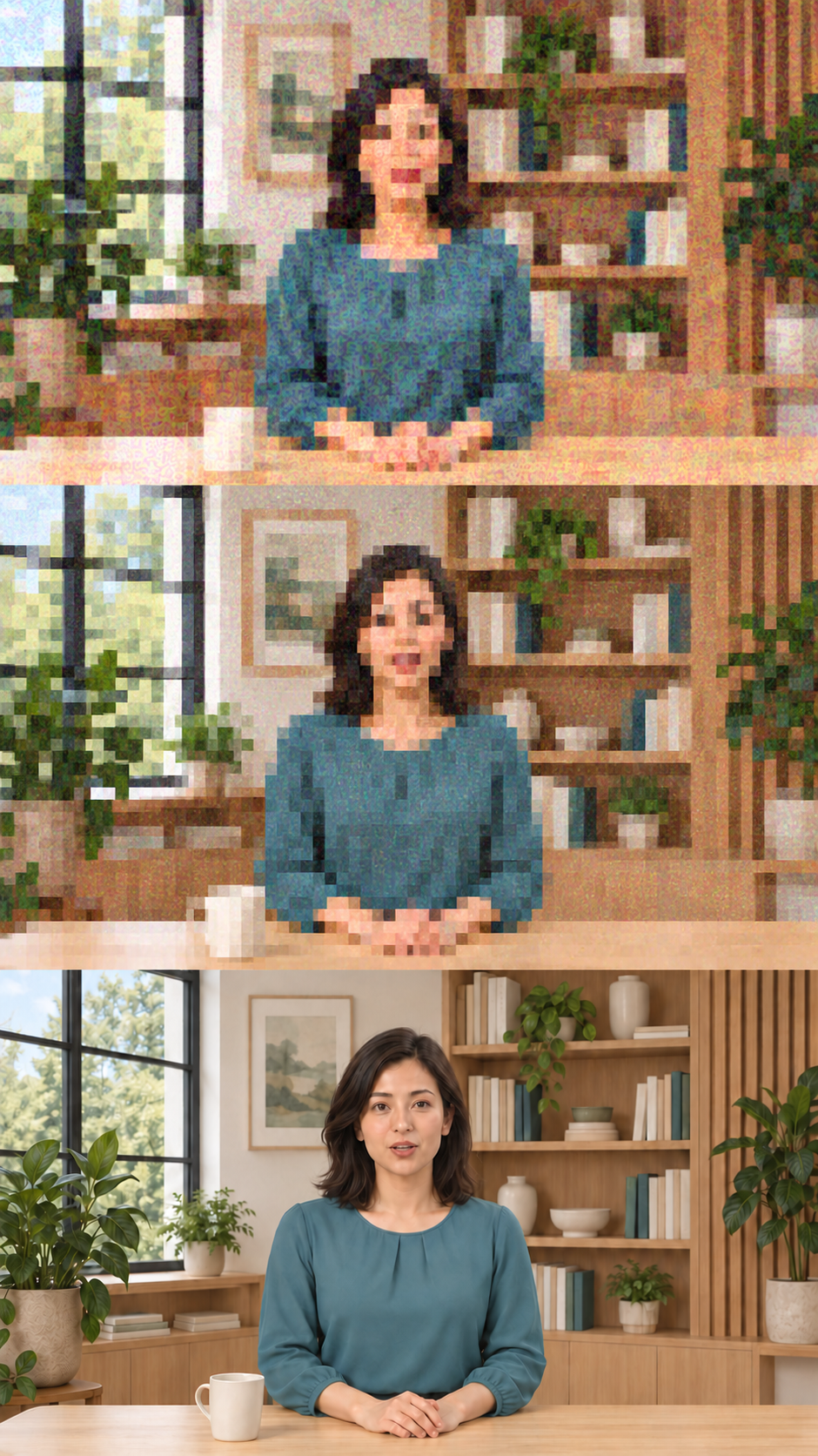}};
  \end{scope}
}
\node[smalltext,text=figmuted] at (21,32.7) {$15\!\times\!27$ video grid};
\node[smalltext,text=figmuted] at (69,32.7) {$15\!\times\!27$ video grid};
\node[smalltext,text=figmuted] at (118,32.7) {$24\!\times\!43$ video grid};
\node[smalltext] at (21,35.2) {Clean estimate $\widehat x^v_{b,1}$};
\node[smalltext] at (69,35.2) {Clean estimate $\widehat x^v_{b,2}$};
\node[smalltext] at (118,35.2) {Clean endpoint $\widehat x^v_{b,3}$};
\draw[flow] (low0.east) -- (low16.west);
\node[smalltext,text=figmuted] at (45,20.6) {Euler};
\draw[flow] (low16.east) -- (high33.west);
\node[smalltext,text=fighigh] at (93.5,19.6) {Euler +\\coupled lift};

\node[smalltext,text=figmuted] at (7.5,39.6) {Audio};
\foreach \xx/\jj in {21/1,69/2,118/3}{
  \draw[condition] (\xx,37) -- (\xx,38.7);
  \foreach \token in {0,...,5}{
    \filldraw[fill=figmem!40,draw=figmem!75,line width=.25pt,rounded corners=.2mm]
      ({\xx-6+2*\token},38.8) rectangle ++(1.6,1.8);
  }
  \node[bank] (kv\jj) at (\xx,45.6) {Clean AV $\rightarrow\mathcal K_{b,\jj}$};
  \draw[condition] (\xx,40.6) -- (\xx,42.8);
  \node[next,text=figlow] (next\jj) at (\xx,52.4) {Stage $\jj$, chunk $b+1$};
  \draw[history] (kv\jj.south) -- (next\jj.north);
}
\draw[condition] (27,39.7) -- (62.5,39.7);
\draw[condition] (75,39.7) -- (111.5,39.7);

\fill[figlow!10,rounded corners=1.3mm] (0,58) rectangle (98,81.7);
\fill[figmem!12,rounded corners=1.3mm] (101,58) rectangle (139.7,81.7);
\draw[figlow,line width=.9pt] (0,58) -- (98,58);
\draw[figmem,line width=.9pt] (101,58) -- (139.7,58);
\node[heading] at (2,61) {Asymmetric distillation};
\node[heading,font=\sffamily\fontsize{7.5}{8.5}\selectfont\bfseries]
  at (103,61) {Stage-parallel inference};

\node[tag=figlow,minimum width=22mm] (student) at (12,65.8)
  {\textbf{Student} lookahead};
\node[tag=figmem,minimum width=18mm] (fake) at (60.5,65.8) {\textbf{Fake} initial};
\node[tag=fighigh,minimum width=19mm] (real) at (82,65.8) {\textbf{Real} original};
\node[box=figlow,minimum width=22mm] (exit) at (12,73.1) {Selected stage\\clean prediction};
\node[box=fighigh,minimum width=22mm] (canvas) at (39.5,73.1) {Shared HIGH\\AV canvas};
\node[box=figmem,minimum width=25mm] (scores) at (67.5,73.1) {Real / Fake\\same noise, $\sigma$};
\node[box=figink,minimum width=12mm,fill=black!13] (dmd) at (90.5,73.1) {DMD};
\draw[condition] (student.south) -- (exit.north);
\draw[condition] (fake.south) -- (60.5,69.6);
\draw[condition] (real.south) -- (75,69.6);
\draw[flow] (exit.east) -- (canvas.west);
\draw[flow] (canvas.east) -- (scores.west);
\draw[flow] (scores.east) -- (dmd.west);
\draw[history] (dmd.south) -- (90.5,79.8) -- (12,79.8) -- (exit.south);
\node[smalltext,fill=figlow!10,inner sep=.5pt,text=figmem]
  at (50,79.8) {Gradient through clean resize};

\foreach \row/\jj/\accent in {0/1/figlow,1/2/figlow,2/3/fighigh}{
  \node[smalltext,anchor=east,text=figmuted] at (106.7,{66+5*\row}) {$\jj$};
  \foreach \chunk/\chunklabel in {0/b,1/{b\!+\!1},2/{b\!+\!2}}{
    \pgfmathsetmacro{\xx}{108+6*(\row+\chunk)}
    \pgfmathsetmacro{\yy}{64+5*\row}
    \filldraw[fill=\accent!28,draw=\accent!75,line width=.35pt,rounded corners=.55mm]
      (\xx,\yy) rectangle ++(5.6,4.1);
    \node[smalltext,text=black] at ({\xx+2.8},{\yy+2.05}) {$\chunklabel$};
  }
}
\draw[condition] (108,80) -- (137,80);
\end{tikzpicture}
}
\endgroup
\caption{\textbf{A chunk's refinement becomes the next chunk's history.}
The $N=3$ configuration shown here uses two coarse-grid evaluations followed by high-resolution refinement.
Each stage estimates a clean AV endpoint and encodes it on its native grid for the same stage of the next chunk, enabling the parallel execution shown at lower right.
Audio keeps its latent geometry throughout.
The lower-left panel shows training: selected-stage predictions form a high-resolution sequence for the two distillation scores.
The generator and scores share scene and speaker conditions, with role-specific dialogue windows.}
\label{fig:method_pipeline}
\end{figure*}

%% file: sections/4_experiments.tex
\section{Experiments}
\label{sec:experiments}

\begin{table}[tbp]
\caption{Long-form quality on the 100-input benchmark and single-stream efficiency.
Serial and parallel configurations are reported separately, with timing on eight H20 GPUs.
Bold marks the best available quality score within the streaming audio-video group only.
Published single-GPU rates are from \citet{taomate2026}.
VisualRealism, AudioAes, LipSync, and Desync use VABench.
S2V LipSync and Desync use the conditioning speech, while AudioAes and CER are not scored as generation metrics.
CER excludes methods without Chinese speech support.
}
\label{tab:cross_method}
\centering
\small
\setlength{\tabcolsep}{3.0pt}
\renewcommand{\arraystretch}{0.90}
\resizebox{\linewidth}{!}{%
\begin{tabular}{@{}lccccccccc@{}}
\toprule
& \multicolumn{2}{c}{Continuity proxies$\uparrow$} & \multicolumn{2}{c}{AV quality$\uparrow$} & \multicolumn{3}{c}{Speech / sync} & \multicolumn{2}{c}{Efficiency}\\
\cmidrule(lr){2-3}\cmidrule(lr){4-5}\cmidrule(lr){6-8}\cmidrule(l){9-10}
Method & Temporal & \shortstack{Long-\\range} & \shortstack{Visual\\Realism} & AudioAes & \shortstack{CER (\%)\\$\downarrow$} & \shortstack{LipSync\\$\uparrow$} & \shortstack{Desync\\$\downarrow$} & GPUs & FPS$\uparrow$\\
\midrule
\multicolumn{10}{@{}l}{\emph{Audio-video foundation and accelerated clip models}}\\
MiniMax-H3~\citep{minimaxh3_2026} & 0.872 & 0.673 & 5.00 & 4.335 & 3.42 & 7.569 & 0.000 & 8 & 0.40\\
VDN~\citep{xi2026videodeltanet} & 0.866 & 0.750 & 5.00 & 4.264 & 3.59 & 7.844 & 0.000 & 8 & 2.21\\
LTX-2.3~\citep{ltx2_2026} & 0.544 & 0.453 & 2.65 & 3.849 & 14.51 & 4.249 & 0.050 & 1 & 0.301\\
JavisDiT++~\citep{javisditpp2026} & 0.095 & 0.423 & 4.95 & 3.068 & -- & 0.899 & 1.050 & 1 & 0.477\\
MOVA~\citep{mova2026} & 0.184 & 0.821 & 4.10 & 4.317 & 87.58 & 3.993 & 0.060 & 1 & 0.145\\
OVI~\citep{ovi2025} & 0.773 & 0.842 & 3.40 & 3.347 & 72.65 & 3.622 & 0.120 & 1 & 0.278\\
\midrule
\multicolumn{10}{@{}l}{\emph{Speech-to-video (S2V)}}\\
OmniAvatar~\citep{omniavatar2025} & 0.827 & 0.721 & 4.80 & -- & -- & 4.859 & 0.010 & 1 & 1.022\\
StableAvatar~\citep{stableavatar2025} & 0.835 & 0.896 & 5.00 & -- & -- & 3.633 & 0.130 & 1 & 0.409\\
Wan2.2-S2V~\citep{wans2v2025} & 0.860 & 0.885 & 4.95 & -- & -- & 5.030 & 0.150 & 1 & 0.080\\
LiveAvatar~\citep{liveavatar2025} & 0.874 & 0.969 & 4.90 & -- & -- & 5.425 & 0.180 & 1 & 4.800\\
Hallo3~\citep{hallo3_2025} & 0.784 & 0.782 & 4.95 & -- & -- & 4.798 & 0.010 & 1 & 0.169\\
\midrule
\multicolumn{10}{@{}l}{\emph{Streaming audio-video generation}}\\
OmniForcing~\citep{omniforcing2026} & 0.786 & 0.278 & 4.85 & 3.571 & -- & 1.756 & 0.050 & 1 & 8.32\\
Lynn~\citep{mao2026lynnrealomninativemultimodalvideo} & 0.367 & 0.778 & 4.30 & 4.207 & 86.70 & \textbf{7.254} & 0.030 & 1 & 1.05\\
Lynn-Flash~\citep{mao2026lynnrealomninativemultimodalvideo} & 0.457 & 0.765 & \textbf{5.00} & 4.326 & 88.37 & 5.909 & \textbf{0.000} & 1 & 1.66\\
TaoMate~\citep{taomate2026} & 0.843 & \textbf{0.983} & 4.65 & 3.889 & 103.74 & 5.422 & 0.010 & 1 & 11.1\\
\midrule
\method (serial) & \textbf{0.913} & 0.930 & \textbf{5.00} & \textbf{4.407} & \textbf{19.57} & 7.061 & \textbf{0.000} & 8 & 4.68\\
\method (parallel) & \textbf{0.913} & 0.928 & \textbf{5.00} & 4.367 & 21.28 & 6.724 & \textbf{0.000} & 8 & 8.99\\
\bottomrule
\end{tabular}}
\end{table}

\subsection{Evaluation setup}
\label{sec:eval_setup}

We evaluate long-form quality, isolate spatial-schedule efficiency, and measure gains from deployment and continuation training.

\paragraph{Model and sequences.}
We train rank-128 LoRA on 10,000 self-constructed distillation pairs, starting from MiniMax-H3 FL2VA model \citep{minimaxh3_2026}.
We evaluate the methods on 100 self-constructed sequences lasting 1--10 minutes, each comprising successive five-second text requests.
Outputs are \highcanvas{} at 24 FPS with 32-kHz stereo audio.
We also generate hour-long audiovisual streams by continuing the autoregressive rollout across successive text windows.
Figure~\ref{fig:teaser} shows samples at $1376\times768$, with frames sampled from the opening to 60 seconds to illustrate persistent character appearance and scene composition.

\paragraph{Quality and efficiency metrics.}
Table~\ref{tab:cross_method} uses VABench VisualRealism, AudioAes, LipSync, and Desync for visual quality, audio quality, and audiovisual synchronization \citep{vabench2026}.
Temporal measures frame-difference stability, while Long-range measures color/brightness continuity across full sequences.
CER compares Whisper-small transcripts \citep{radford2023whisper} with the dialogue.
SyncNet LSE-C/D \citep{chung2017syncnet,prajwal2020wav2lip} average valid five-second windows over each full output.
Mel-frequency cepstral coefficients (MFCCs) summarize the short-time spectral envelope.
We compare their mean-and-standard-deviation vectors over the first and last five seconds by cosine similarity to measure endpoint acoustic consistency.
For the spatial ablation, Video averages normalized sharpness, contrast, exposure, and saturation, while Audio averages loudness, clipping, silence prevalence, and audiovisual-duration agreement.
Efficiency is measured by GPUs per stream, generation FPS, and first ordered RGB latency.

\begin{figure}[!t]
\centering
\includegraphics[width=\linewidth]{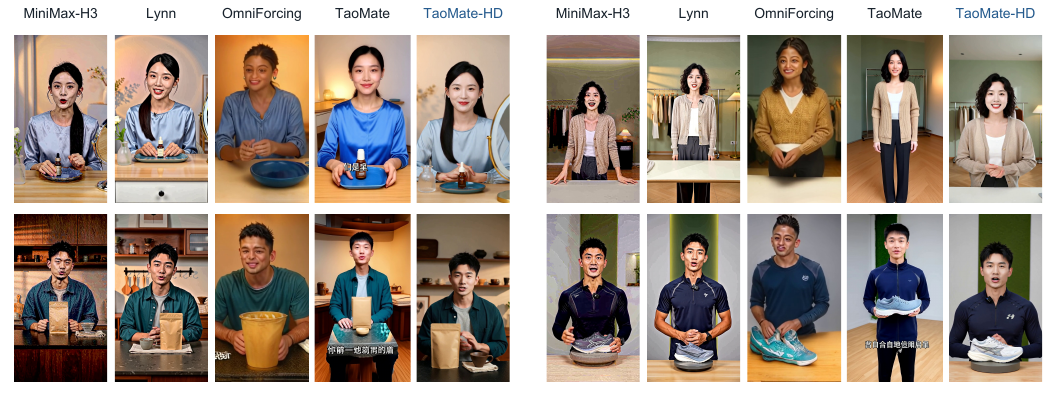}
\caption{Long-form audio-video generation, shown at 60 seconds.
The corresponding quantitative comparison appears in Table~\ref{tab:cross_method}.}
\label{fig:cross_method_main}
\end{figure}

\subsection{Comparison with audio-video generators}
\label{sec:cross_method}

Table~\ref{tab:cross_method} organizes the comparison into audio-video foundation and accelerated clip models, speech-to-video (S2V) systems, and streaming audio-video generators.
The first group comprises MiniMax-H3~\citep{minimaxh3_2026}, VDN~\citep{xi2026videodeltanet}, LTX-2.3~\citep{ltx2_2026}, JavisDiT++~\citep{javisditpp2026}, MOVA~\citep{mova2026}, and OVI~\citep{ovi2025}.
The S2V group comprises OmniAvatar~\citep{omniavatar2025}, StableAvatar~\citep{stableavatar2025}, Wan2.2-S2V~\citep{wans2v2025}, LiveAvatar~\citep{liveavatar2025}, and Hallo3~\citep{hallo3_2025}, which animate video from supplied speech.
The streaming group comprises OmniForcing~\citep{omniforcing2026}, Lynn Standard/Flash~\citep{mao2026lynnrealomninativemultimodalvideo}, and TaoMate~\citep{taomate2026}, which jointly generate speech and video across successive temporal blocks.

The models extend generation through different forms of context.
MiniMax-H3 and VDN extend an opening T2AV clip with successive I2AV clips.
Lynn retains persistent media KV, while ours combines retained audiovisual history with overlapping dialogue and evolving scenes.
MiniMax-H3 uses 49 denoising evaluations per five-second request, VDN uses eight, and the Lynn variants use four or three per autoregressive window.
Our model uses three denoising evaluations and three clean-history encodings per chunk.

Table~\ref{tab:cross_method} reports quality for both serial and parallel configurations.
Our interval spans prepared input to decoded AV, while the MiniMax-H3, VDN, and Lynn intervals include online conditioning and encoding/mux.
All published rates measure denoising on one or eight NVIDIA H20 at $768\times512$ \citep{taomate2026}.
Under the respective timing intervals above, the reported throughput ratio to MiniMax-H3 is $22.4\times$.

\paragraph{Comparison within streaming audio-video generation.}
Serial \method has the highest Temporal and AudioAes scores in Table~\ref{tab:cross_method}: 0.913 and 4.407, compared with 0.786 and 3.571 for OmniForcing and 0.843 and 3.889 for TaoMate.
It achieves the lowest reported CER in this group at 19.57\%.
It also shares the best VisualRealism and Desync values with Lynn-Flash.
The parallel configuration retains the same rounded Temporal and VisualRealism scores, with AudioAes of 4.367 and CER of 21.28\%.
TaoMate has the highest long-range color/brightness score.

\paragraph{Visual comparison.}
Figure~\ref{fig:cross_method_main} complements the sequence-level measurements with corresponding outputs at 60 seconds.
Presenter appearance, hand and product geometry, and background texture show the visual detail near the end of each sequence.

\FloatBarrier
\subsection{Spatial efficiency and deployment throughput}
\label{sec:spatial_ablation}

\begin{figure}[htbp]
\centering
\includegraphics[width=0.99\linewidth]{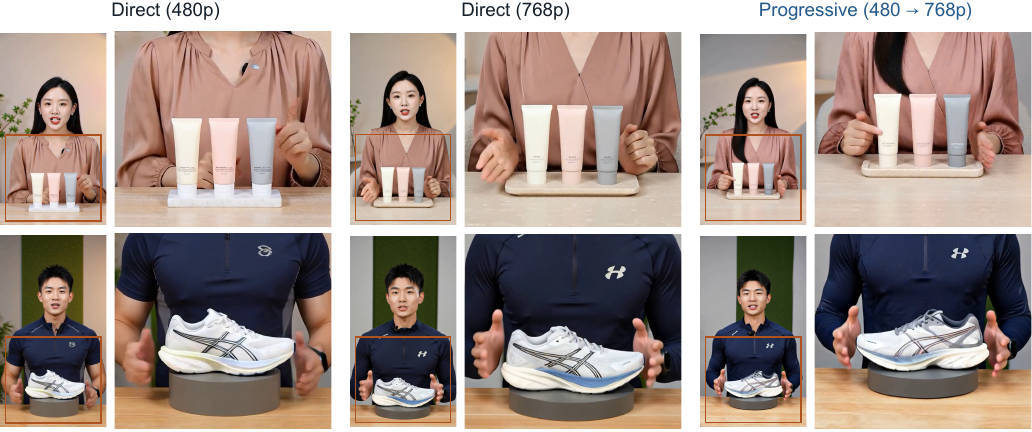}
\caption{\textbf{Spatial detail across generation resolutions.}
Representative frames compare direct 480p, direct 768p, and progressive 480-to-768p generation on the same inputs.
Enlarged foreground regions reveal hand contours and product surfaces.
Progressive generation retains fine detail comparable to direct 768p generation while concentrating denoising on the coarse grid.}
\label{fig:spatial_qualitative}
\end{figure}

\begin{table}[htbp]
\caption{Resolution, quality, and inference throughput.
All experiments are conducted on the same batch of input settings and GPUs.
Direct (480p) is a separate low-resolution reference, and \method is the stage-parallel configuration.}
\label{tab:spatial_ablation}
\centering
\small
\setlength{\tabcolsep}{5pt}
\begin{tabular}{@{}lrrrrrr@{}}
\toprule
Configuration & FPS$\uparrow$ & Temporal$\uparrow$ & \shortstack{Long-range$\uparrow$} & Video$\uparrow$ & Audio$\uparrow$ & LSE-C$\uparrow$\\
\midrule
Direct (480p) & 13.90 & 0.90 & 0.99 & 0.88 & 0.85 & 5.98\\
Direct (768p) & 2.93 & 0.91 & 0.92 & 0.87 & 0.89 & 6.62\\
Progressive (serial) & 4.68 & 0.91 & 0.93 & 0.85 & 0.89 & 6.34\\
\midrule
\method & 8.99 & 0.91 & 0.93 & 0.85 & 0.89 & 6.39\\
\bottomrule
\end{tabular}
\end{table}

Table~\ref{tab:spatial_ablation} compares resolution and execution settings, with direct 480p generation as the low-resolution reference.
The middle FPS entries compare spatial schedules, while the final row reports deployment throughput.
For the spatial timing comparison, Direct (768p) and serial Progressive share the generator, inputs, request seeds, stage-local AV history, and BF16/SDPA execution, changing the resolution schedule while retaining the temporal schedule and decoding policy.
Timing includes generation and audiovisual decoding after warmup and text preencoding.

Averaging wall time equally over inputs and repeated executions, Progressive reduces generation time from 491 to 307 seconds.
The ratio of these mean times gives a $1.60\times$ spatial speedup (Table~\ref{tab:spatial_ablation}).
The analytic video-linear work ratio is 0.595, whereas measured wall time additionally includes context encoding, audio computation, communication, and decoding.
Figure~\ref{fig:spatial_qualitative} shows the detail retained under this reduced spatial workload: enlarged views expose hand contours and product surfaces after progressive refinement to 768p.

We next combine the progressive trajectory with the optimized stage-parallel deployment.
Mean generation time decreases from 307 seconds for serial progressive to 160 seconds, a $1.92\times$ overall acceleration (Table~\ref{tab:spatial_ablation}).
This is a deployment-level comparison: it combines overlapping stages, low-stage fusion, KV reuse, selective W8A8 projections, a different attention implementation, and codec scheduling.
The resulting configuration reaches 8.99 FPS, with its quality measurements listed in the final row.

\FloatBarrier
\subsection{Ablation study of asymmetric distillation}
\label{sec:speech_ablation}

\begin{figure}[htbp]
\centering
\includegraphics[width=\linewidth,trim=0 22 0 0,clip]{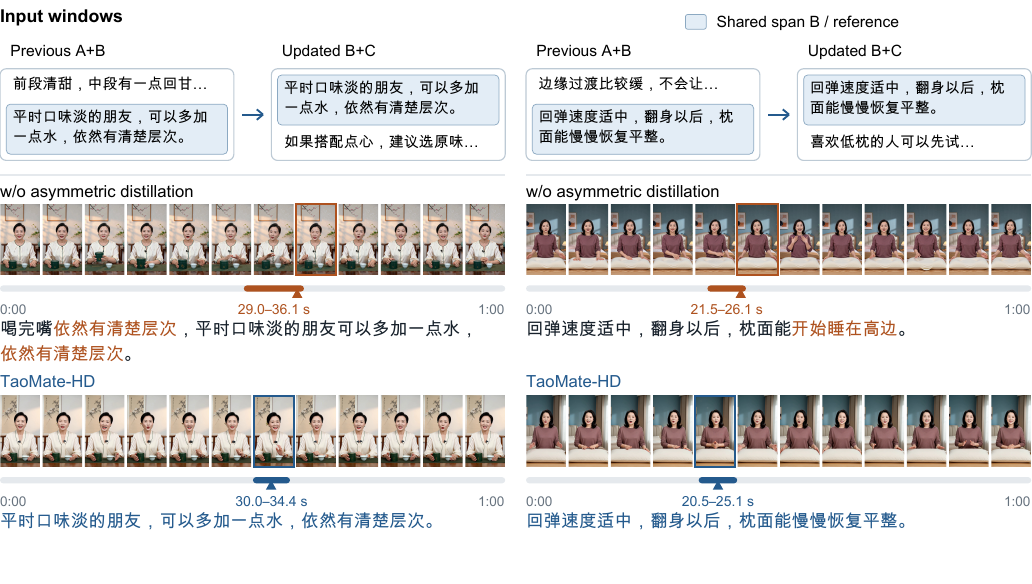}
\par\captionsetup{skip=3pt}
\caption{\textbf{Text windows and generated speech.}
The input strip shows the overlapping windows supplied to \method, with the shared span $B$ as the reference excerpt.
Thumbnail strips span each video, and colored intervals mark the transcribed excerpts.
Its transcripts show a repeated clause (left) and an incorrect sentence continuation (right).
Orange marks the affected phrases.}
\label{fig:recipe_speech_qualitative}
\end{figure}

The proposed asymmetric distillation reduces CER from 30.8\% to 19.6\%, an 11.2-point improvement (Table~\ref{tab:speech_breakdown}).
The two continuation-training configurations share the fixed-resolution initialization and progressive generation, with different text conditioning, history construction, rollout settings, augmentation, and auxiliary supervision.
The configuration labeled \continuationbaseline shares text between the generator and both scores.
Endpoint MFCC cosine increases from 0.971 to 0.979, indicating more consistent acoustic characteristics between the opening and closing speech.
LSE-C changes from 6.239 to 6.341 and LSE-D from 9.011 to 9.044.

\begin{table}[H]
\caption{Comparison of distillation training configurations.
CER is the mean character-error percentage.
MFCC cosine measures endpoint acoustic consistency between the first and last five seconds.}
\label{tab:speech_breakdown}
\centering
\small
\setlength{\tabcolsep}{8pt}
\begin{tabular}{@{}lcccc@{}}
\toprule
Configuration & \shortstack{CER (\%)$\downarrow$} & \shortstack{MFCC cosine$\uparrow$} & LSE-C$\uparrow$ & LSE-D$\downarrow$\\
\midrule
\shortstack[l]{\textit{w/o asymmetric}\\\textit{distillation}} & 30.751 & 0.97064 & 6.239 & \textbf{9.011}\\
\method & \textbf{19.570} & \textbf{0.97905} & \textbf{6.341} & 9.044\\
\bottomrule
\end{tabular}
\end{table}

The CER breakdown shows decreases of 12.64 and 1.28 percentage points in substitution and deletion errors, alongside a 2.74-point increase in insertion errors.

Figure~\ref{fig:recipe_speech_qualitative} illustrates deviations from the intended text using \method and a baseline training without asymmetric distillation.
The displayed input windows identify the shared words retained when \method advances its text condition.
The baseline repeats an upcoming clause or substitutes the intended sentence ending with words from the following sentence.

%% file: sections/5_conclusion.tex
\Needspace{4\baselineskip}
\section{Conclusion}
\label{sec:conclusion}

We presented \method to reduce the computational cost of high-resolution audiovisual streaming while maintaining speech continuity.
Progressive denoising establishes coarse video structure before final high-resolution refinement, while stage-local KV history enables overlapping generation across chunks.
Distillation with asymmetric text conditioning teaches the model to use retained audiovisual history as dialogue windows advance.
Experiments show that progressive generation achieves a $1.60\times$ speedup over direct generation while retaining fine visual detail.
The optimized stage-parallel deployment provides a further $1.92\times$ speedup. Asymmetric distillation reduces CER from 30.8\% to 19.6\%.
Together, these advances support high-resolution audiovisual generation extending to an hour.